\documentclass[sigconf]{acmart}

\usepackage{booktabs} % For formal tables
\usepackage[german,american]{babel}
\usepackage[T1]{fontenc}
\usepackage[latin1]{inputenc} % For umlauts (OT1) and the Ã (OT1, T1).
\usepackage{array}
\usepackage{enumitem}
\usepackage{todonotes}
\usepackage{amsmath}

\DeclareGraphicsExtensions{.pdf,.ai,.jpg,.png}
\setkeys{Gin}{pagebox=artbox}% Alternative for \pdfpagebox 5 in preceding line.
\graphicspath{{../chiir19-comparative-argumentative-machine-figures/}}
\usepackage{tabularx}

\begin{document}

\title[Answering Comparative Questions: Better than Ten-Blue-Links?]{Answering Comparative Questions: Better than Ten-Blue-Links?}
%\titlenote{We can add a note to the title}
\copyrightyear{2019} 
\acmYear{2019} 
\setcopyright{rightsretained} 
\acmConference[CHIIR '19]{2019 Conference on Human Information Interaction and Retrieval}{March 10--14, 2019}{Glasgow, United Kingdom}
\acmBooktitle{2019 Conference on Human Information Interaction and Retrieval (CHIIR '19), March 10--14, 2019, Glasgow, United Kingdom}
\acmDOI{10.1145/3295750.3298916}
\acmISBN{978-1-4503-6025-8/19/03}

%\author{Matthias Schildw{\"a}chter$^\ddag$, Alexander Bondarenko$^\dag$, Julian Zenker$^\ddag$, Matthias Hagen$^\dag$,\\ \vspace{-10pt} Chris Biemann$^\ddag$, and Alexander Panchenko$^\ddag$}

%\affiliation{University of Hamburg}

%\email{{7schildw,panchenko,6zenker,biemann}@informatik.uni-hamburg.de}
%\email{{alexander.bondarenko,matthias.hagen}@informatik.uni-halle.de}

\author{Matthias Schildw{\"a}chter}
\affiliation{University of Hamburg, Germany}
%\email{7schildw@informatik.uni-hamburg.de}

\author{Alexander Bondarenko}
\affiliation{Martin Luther University of Halle-Wittenberg, Germany}
%\email{alexander.bondarenko@informatik.uni-halle.de}

\author{Julian Zenker}
\affiliation{University of Hamburg, Germany}
%\email{6zenker@informatik.uni-hamburg.de}

\author{Matthias Hagen}
\affiliation{Martin Luther University of Halle-Wittenberg, Germany}
%\email{matthias.hagen@informatik.uni-halle.de}

\author{Chris Biemann}
\affiliation{University of Hamburg, Germany}
%\email{biemann@informatik.uni-hamburg.de}

\author{Alexander Panchenko}
\affiliation{University of Hamburg, Germany}
%\email{panchenko@informatik.uni-hamburg.de}

\begin{abstract}

We present CAM (comparative argumentative machine), a novel open-domain IR~system to argumentatively compare objects with respect to information extracted from the Common~Crawl. In a user study, the participants obtained 15\%~more accurate answers using~CAM compared to a ``traditional'' keyword-based search and were 20\%~faster in finding the answer to comparative questions.
\end{abstract}


%\ccsdesc[300]{Computer systems organization~Graphical User Interfaces}
% \ccsdesc{}



\maketitle

\renewcommand{\shortauthors}{}

\section{Introduction}

Everyone faces choice problems on a daily basis. Besides choosing what to wear or what to have for lunch, people compare all kinds of options: cameras to buy, universities to study at, or even programming languages to use. Question answering  platforms like Quora, Reddit, or StackExchange are packed with comparative questions like ``How does X compare to Y with respect to Z?''. An informed choice then is often based on an objective argumentation why to favor one of the candidates (e.g., comparing important aspects).

Specific product comparison systems, such as Compare.com or Check24, allow to compare any subset of objects in narrow domains such as cameras. Other systems like WolframAlpha aim at providing comparative functionality across domains, but also often only use some (limited) structured database while ignoring the rich textual content available on the web. Somewhat surprising, no system is currently able to satisfy comparative information needs for the general domain with sufficient coverage and explanations. No available system is able to support comparisons on a broad range of object types with arguments about relative qualities or even supporting objective arguments about the best choice. Indeed, web search engines are able to directly answer many factoid questions but do not treat comparative questions any special beyond returning default search results. Advanced question answering systems, such as IBM's Watson~\cite{ferrucci2010building}, answer factoid questions very well, but do not really handle comparative questions of everyday users.

We present CAM (comparative argumentative machine), a system that aims at solving the shortcomings mentioned above. CAM is a tool for answering comparative general-domain questions based on information extracted from the web-scale Common Crawl.\footnote{\textbf{Demo, API \& code: \url{http://ltdemos.informatik.uni-hamburg.de/cam/}}}

\section{Related Work}

Commercial systems like GoCompare, Compare.com, Diffen.com, and Versus.com offer high-precision comparison capabilities based on well-curated structured data sources focusing on single domains. But their low coverage in other than their focus domains rules out answering most of the comparative questions found on portals like Quora or Yahoo!\,Answers---that themselves form a good source of (argumentative) comparisons and results.

Previous text-based comparison approaches have mostly focused on the biomedical domain. \citet{fiszman2007interpreting} collected sentences comparing drug therapies using manually crafted patterns to recognize the subjects of comparison and the comparison direction. They reached a very high precision at moderate recall. On a set of full-text articles on toxicology, \citet{park2012identifying} succeeded in training a highly precise Bayesian Network for identifying comparative sentences relying on lexical clues and dependency parsers. More recently, \citet{gupta2017identifying} described a system based on manually collected patterns on the basis of lexical matches and dependency parses in order to identify comparison targets and to classify the type of comparison into the classes given by \citet{JindalLiu2006}: gradable vs.\ non-gradable and superlative comparisons. 

Building a general-domain argumentative comparison facility comes with the additional challenge of argument mining from user-generated content~\cite{Snajder2017Social-Media-Ar}. Text is typically noisy, misses argument structures and contains poorly formulated claims. On the other hand, specialized jargon and idiosyncrasies of a platform can be utilized~\cite{Dusmanu2017Argument-Mining} (e.g., hashtags for mining argumentative tweets).

\citet{Aker2017What-works-and-} confirmed the findings of \citet{Stab2014Identifying-Arg} that information from dependency parsers does not help to find the (general) argument structure in persuasive essays and Wikipedia articles while simpler structural features such as punctuation are more effective. \citet{Daxenberger2017What-is-the-Ess} noted that claims across different domains share lexical clues and further stated that current datasets are too small for recent DNN-based classifiers resorting to traditional feature engineering for argument mining. 

Some argument mining systems work on larger corpora of user-generated content to find the most relevant argument for a given claim~\cite{hua-wang:2017:Short} or to oppose different argumentative viewpoints~\cite{wachsmuth2017building}. Web-scale systems for comparing query results~\cite{sun2006cws} or for retrieving single arguments matching a user query~\cite{stab2018argumentext} form the inspiration for our new CAM~system (comparative argumentative machine). 

\section{The CAM System Design}

To ensure a wide coverage, a comparative answer of our CAM~system for two objects is based on argumentative structures extracted from web-scale text resources. The system looks for textual structures asserting that one of the compared objects is superior to the other, that they are equal, or that they are not comparable. A comparison of two objects~$o$ and~$o'$ in the CAM~sense is defined as ``$o \text{ \textbf{?} } o' \text{ w.r.t. } a_i, \dots, a_j \in \mathbf{a}$'', where \textbf{?} is in $\{>, <, =, \neq\}$ and $\mathbf{a} = \{a_1, \ldots, a_k\}$ is the set of comparison aspects of~$o$ and~$o'$. We thus focus on mining sentences like ``\textit{Python}=${o}$ is better than \textit{Matlab}=${o'}$ for \textit{web development}=${a_i}$.''

\begin{figure}[t]
    \centering
    \includegraphics[width=0.47\textwidth]{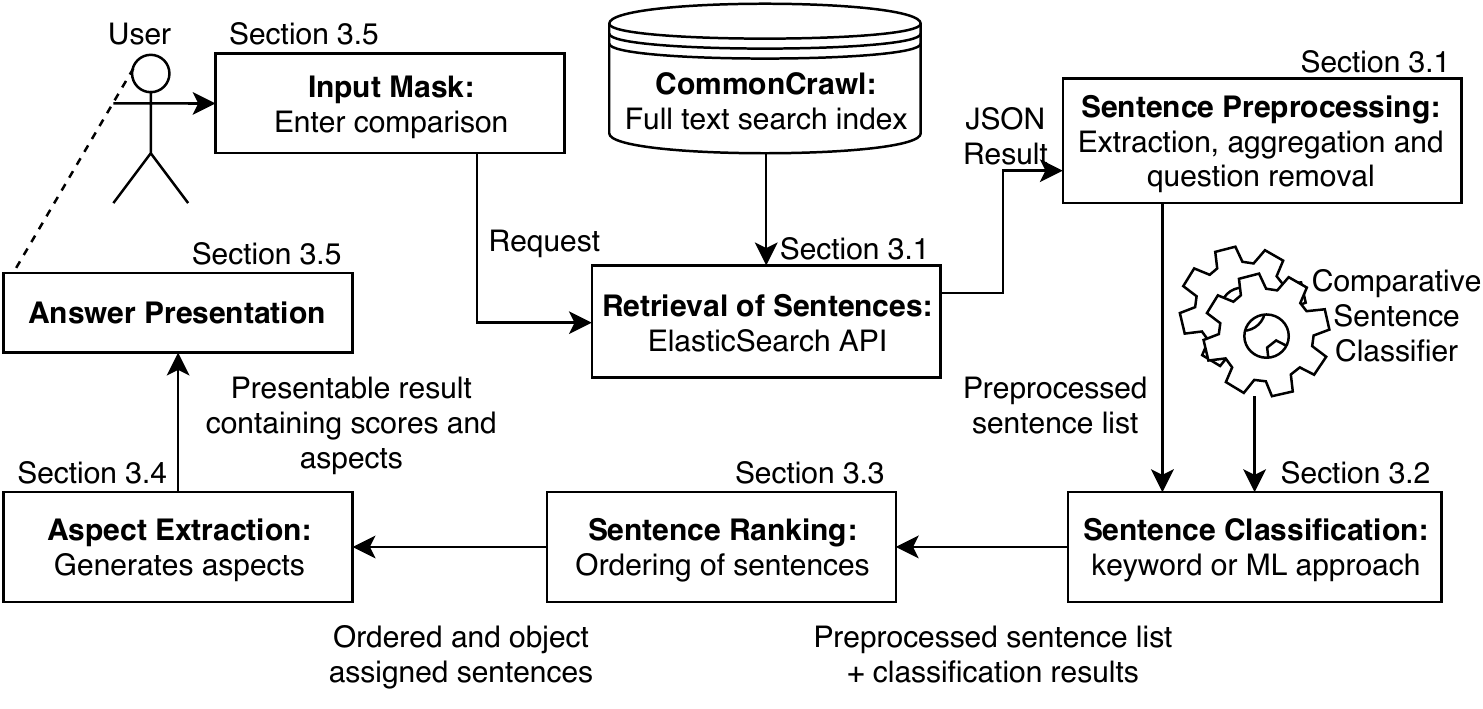}
    \caption{Design of the CAM~system.}
    \label{fig:pipeline}
\end{figure}

The design of our CAM~system is shown in Figure~\ref{fig:pipeline}. It consists of the following generic stages, which are further described in details: (1)~retrieval of relevant sentences, (2)~classification of comparative sentences, (3)~ranking of the comparative sentences, (4)~extraction of object aspects, and (5)~presentation of the answer.

\subsection{Sentence Retrieval}

Our CAM system uses an Elasticsearch full text index of a linguistically pre-processed corpus~\cite{PANCHENKO18.215} containing 14.3~billion English sentences from the Common~Crawl. To retrieve textual argumentative structures relevant to a comparative user input, the index is queried for sentences matching the input objects and containing  comparison aspects; sentences without aspects are used as a fall-back. Questions are removed from the initial retrieval results since they usually do not help in returning an argumentative answer.

\begin{figure*}[t]
\includegraphics[width=0.88\linewidth]{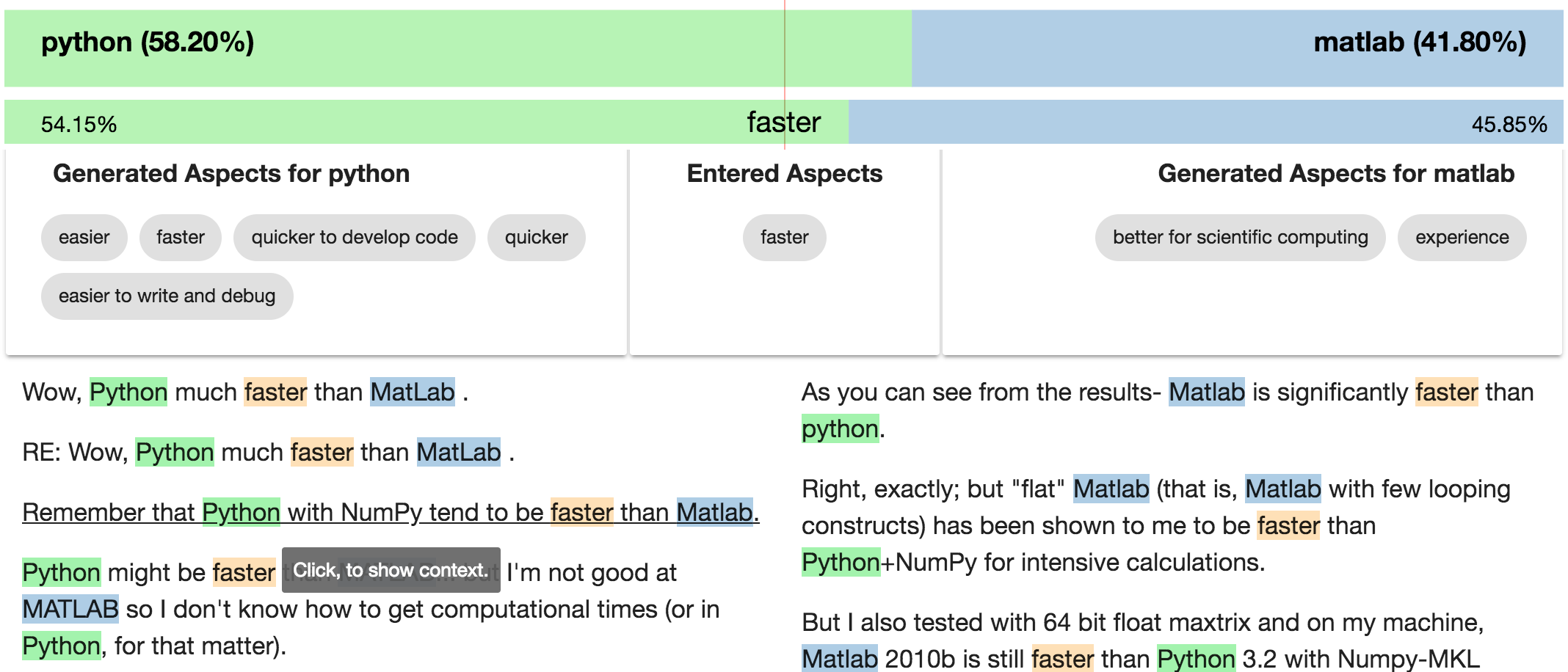}
\caption{CAM answer presentation for the question ``Is Python faster than Matlab?''. Pro and con sentences are shown.}
\label{fig:cam-answer-presentation}
\end{figure*}

\begin{figure}[t]
\includegraphics[width=\linewidth]{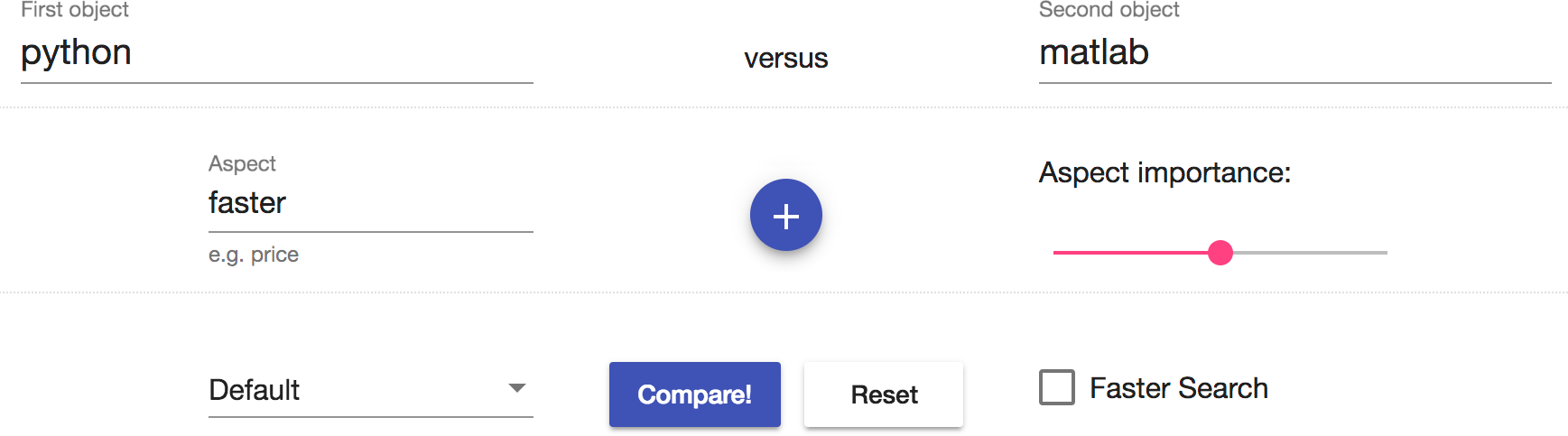}
\caption{CAM input form.}
\label{fig:cam-input-form}
\end{figure}

\subsection{Sentence Classification}

We use a classifier to distinguish between four classes: the first object from the user input is better / equal / worse than the second one ($>, =, <$) w.r.t. a comparison aspect, or no comparison is found~($\neq$). The classifier uses the text between both objects to identify the ``polarity'' and is inspired by the best model reported by \citet{Franzek.9172018}: XGBoost~\cite{chen2016xgboost} using word frequencies as representations, which achieves a high F1~score of~0.92 for~$\neq$, a good F1 of~0.74 for~$>$ but a rather low F1 of~0.46 for~$<$. We identified the main issue in missing negation handling, for which we added a simple  keyword-based heuristic to our CAM~system inverting common  negations.

\subsection{Sentence Ranking and Object Comparison}

To rank comparative sentences (category $>$ or $<$), we score them by combining the classifier confidence and the Elasticsearch score\footnote{\url{https://www.elastic.co/guide/en/elasticsearch/guide/current/scoring-theory.html}} according to the following heuristic~$s$:
\begin{equation*}
  s =\left\{
  \begin{array}{@{}ll@{}}
    \phantom{(}\alpha + e + e_{max}, & \text{if confidence} \geq \gamma, \\
    (\alpha +  e) \cdot \delta  , & \text{otherwise,}
  \end{array}\right.
\end{equation*} 

where $e$ is the Elasticsearch score of the sentence, $e_{max}$ is the maximum Elasticsearch score of any comparative sentence retrieved for the user input, and $\alpha = w_{a_i} e_{max}$ if the user-specified aspect $a_i$ is present in the sentence and $\alpha = 0$ otherwise. For the $\alpha$ aspect boost, the weights $w_{a_i}$ are specifiable in the user interface. Confidently classified sentences obtain a boost of $e_{max}$ while scores of low confidence sentences are decreased by a factor of $\delta$; we set $\gamma = 0.8$ and $\delta=0.1$ in our experiments.

For scoring a CAM~output ``$o > o' \text{ w.r.t. } \mathbf{a}$'', we sum up the $s$-scores of all sentences \emph{supporting} the statement. To this end, we have developed a heuristic to include two directions of comparison and thus taking into account that a statement like ``Python is better than Matlab'' (class~$>$) is also supported by ``Matlab is worse than Python'' (class~$<$); important factors being the object ordering and the polarity.

\subsection{Aspect Extraction}

In addition to user-specified comparison aspects, CAM generates up to ten supplementary aspects (even when no comparison aspect at all was provided by the user). We use three different methods for aspect mining: (1)~searching for comparative adjectives and adverbs; (2)~searching for phrases with comparative adjectives / adverbs and a preposition like \textit{to}, \textit{for}, etc. (e.g., ``\textit{quicker to} develop code'' or ``\textit{better for} scientific computing''); (3)~searching for specific hand-crafted patterns like ``because of higher \textit{speed}'', ``since it has more \textit{options}'', ``as we have proven its \textit{resilience}'' or ``reason for this is the \textit{price}''. An extracted aspect is assigned to the object with the higher co-occurrence frequency (cf.\ Figure~\ref{fig:cam-answer-presentation} for examples).

\subsection{CAM User Interface}

The user interface consists of a question input form (Figure~\ref{fig:cam-input-form}) and an answer presentation component (Figure~\ref{fig:cam-answer-presentation}). The input interface allows to submit a comparative question in the form of two compared objects and their aspects. The answer presentation summarizes the sentences retrieved from the Common~Crawl providing  decision-making support for the informed choice.

\subsubsection*{Input Form}

The input form is divided into three parts (cf.~Figure~\ref{fig:cam-input-form}). On the top, the user enters two comparison target objects. In the middle, the interface allows to add an arbitrary number of aspects and assign them a weight indicating their importance (1 to 5; used to boost the scores of the sentences containing the aspect). On the bottom, one of the three different search models can be selected: \textit{Default} is based on keyphrases like ``better than'' or ``faster than'' to find comparative sentences; \textit{BoW} is built upon the word frequency-based XGBoost classifier described above, and \textit{Infersent} uses sentence embeddings. The \textit{Faster Search} option limits the number of queried fall-back sentences to~500 in order to speed up the answer construction.

\subsubsection*{Answer Presentation}

On the top of the comparative answer presentation (cf.~Figure~\ref{fig:cam-answer-presentation}), different score bars are given. The overall score distribution allows the user to grasp a general impression for the entered comparative question while the aspect-specific score bars show the distribution for the individual user-specified aspects.

Additionally, up to ten automatically generated aspects are presented in a clickable manner to allow the user to only display result sentences for such aspects (disjunctive filter interpretation). The user-specified aspects are used on both result sentence sides while the generated aspects only filter the corresponding column.

The objects in the displayed sentences are highlighted with their respective colors from the score bars, while the aspect highlighting uses different colors. Clicking on a result sentence reveals its Common~Crawl context---by default the $\pm3$ sentences around it, with the possibility of expanding to the whole original document.

\section{Evaluation}
\label{sec:evaluation}

We compare our new CAM~system to a keyword-based search in two user studies with 14~and 9~participants on 34~comparison topics.

\subsection{Experimental Setup}

The 34~topics (two compared objects + one aspect) for our studies were created from comparative Quora questions containing the phrase ``better than'' and also being present on comparison pages like Diffen.com and DifferenceBetween.net. For each topic, we manually double-checked that the underlying corpus of our CAM~system and the keyword-based search (i.e., the 14.3~billion Common~Crawl sentences) allows to answer the comparison; we only included topics with at least 20~support sentences. One of the topics for instance has~\textit{mp3} and~\textit{wma} as objects and \textit{compression ratio} as the aspect. Given the ground truth answer \textit{worse}, a study participant should answer that~mp3 is worse than~wma with respect to compression ratio. To clarify potential ambiguities or subjectivities, descriptions for the participants were added to the topics (e.g., to inform a potential music afficionado who might claim a worse compression rate being better since it might come with an improved sound quality). 

In a \textbf{Group~A} setup, we focus on the question whether users are faster answering correctly when using the CAM~system. A G*Power analysis~\cite{faul2007g} did output a required sample size of 272~comparisons to be a able to measure a statistically significant difference in answer times. We thus decided to engage 14~participants on all 34~topics (477~comparisons). Each participant uses both experimental systems alternatingly (CAM / keyword-based search). Since every participant should work on each topic just once with one of the systems but not the other, we randomly split the 34~topics into two groups (one for each system). To avoid any order bias, the topics of each group were presented in random order.

The participants were informed that they should give an answer as quickly as possible; the whole study took about one hour per participant and ended with a questionnaire. We measured the time for different phases (e.g., the time needed to enter a query and the time needed to determine an answer) and the correctness of a participant's answer with respect to manually derived gold labels. 

In a \textbf{Group~B} setup, we focussed on collecting some more ``natural'' feedback using a less forced study environment. We had 9~different participants (not from Group~A) who were allowed to just ``play'' with the systems for any and as many of our 34~comparison topics as they liked. In total, 85~comparisons were performed.

\subsection{Study Participants}

Among the 14~Group~A participants, 9~were male (5~female), 13~indicated 18--24~as their age (1 was in the 25--34~range), 8~participants had an Engineering \& Computer Science background (3~from Arts, Culture \& Entertainment, 1~from Law \& Public Policy, 2~selected ``other'') with 9~having a Bachelor's degree. The participants characterized themselves as having a proficient (nine) or intermediate (five) English level. Seven participants stated to use comparison websites rarely or never (once a year or less), whereas five used them once a month and two even once a week.  

Group~B consisted of five female and four male participants, 1~participant was 13--17~years old, 2~participants fell in the 18--24 age range, 5~in the 25--34~range, and one in~35--44. This group was dominated by an Engineering \& Computer Science background (five out of nine); one from Education, one from Business, one from Arts, Culture \& Entertainment, and one ``other'' background. Four participants already had a Master's degree, two were students, one had a Bachelor's degree, one a doctorate degree, and one selection of ``other''. Six participants rated their English level as proficient and three as intermediate. Five participants stated they used comparison websites rarely or never, whereas two used them once a month and two even once a week or more. 

\subsection{Results and Discussion}

A Shapiro-Wilk test~\cite{shapiro1965analysis} verified the visual assumption of a log-normal distribution ($\alpha = 0.05$) of the different measured times for CAM~usage and keyword-based search. Therefore, t-tests were used to check whether the null hypothesis of same answer determination times or same total times can be rejected.

\begin{figure}[t]
    \includegraphics[width=0.47\textwidth]{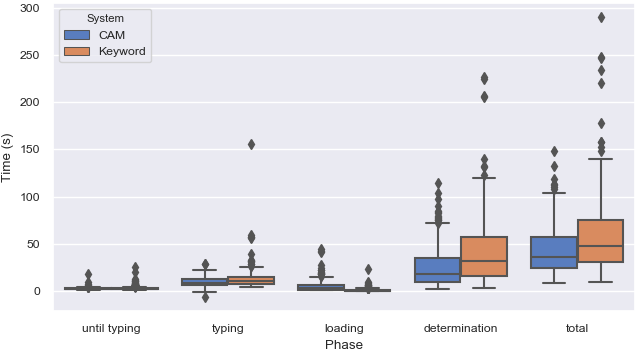}
    \caption{Times of question answering phases (Group A).}
    \label{fig:box-plot-average-results}
\end{figure}

Figure~\ref{fig:box-plot-average-results} shows the time distributions of Group~A. \textbf{Until typing} indicates the participants being about 19\%~faster starting to enter a query with the CAM~system (in Group~B, the CAM~users were even about 25\%~faster). \textbf{Typing} is the time from the first key stroke until the query is submitted. The Group~A participants again were faster with the CAM~interface (about~24\% on average); the Group~B participants needed about twice as long, being slightly faster with the CAM~system. The \textbf{loading} phase measures the time the system needs to show the answer (from sending the query until the result is presented). On average, keyword-based search loads slightly faster than CAM since CAM~uses a keyword-based search subroutine with some further post-processing.

Most importantly, the time the users need to give their answer (\textbf{determination} in Figure~\ref{fig:box-plot-average-results}) shows that the Group~A participants were significantly faster when using the CAM~system (about 39\%~difference). In Group~B, the participants were slower in general, but interestingly they were also slightly slower using~CAM than keyword-based search. One potential reason is that the participants explored the new CAM~interface more even providing verbal feedback during their work (remember that Group~B was allowed to ``play'' with the systems).

For the overall task (\textbf{total} in Figure~\ref{fig:box-plot-average-results}), Group~A was significantly faster when using the CAM~interface while the more exploratory Group~B was overall slower but with no substantial advantage for either system. Our main focus in a Group~B was on observing participants behavior which is why they were allowed to test, play with and comment on the systems while using them.

\begin{figure}[t]
    \includegraphics[width=0.47\textwidth]{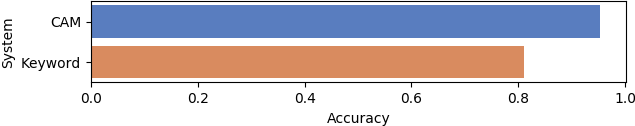}
    \caption{Answer accuracy (Group A).}
    \label{fig:accuracy-summed-h}
\end{figure}

Besides statistically significant quicker answers, the Group~A participants also made fewer errors using the CAM~system (cf.~Figure~\ref{fig:accuracy-summed-h}). The average CAM~accuracy in Group~A is~95\% (9~of 14~participants reached~100\%), whereas for the keyword-search it is~81\% (with a best result of~94\%). The Group~B participants also were more accurate using~CAM (84\%) than using keyword-based search (75\%). 

\begin{figure}[t]
    \includegraphics[width=0.47\textwidth]{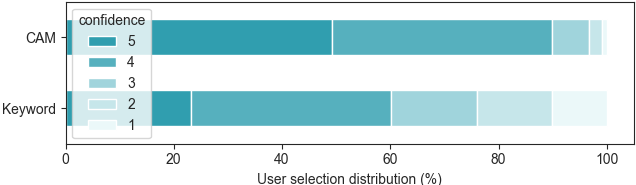}
    \caption{Responses on the question ``How confident are you that the determined answer is correct?'' (Group A).}
    \label{fig:likert-confidence-after-comp}
\end{figure}

In the evaluation questionnaire, we asked the participants to rate the system features on the scale from 1 (very negative) to 5 (very positive).  
The question ``How convenient was it to use the CAM system?'' and the statement ``Learning the usage of CAM is...'' achieved values between~4 and~5 for both groups, which is very positive. In addition, the participants of both groups on average were almost one point more confident that an answer determined by CAM was correct than for keyword-based search (cf.~Figure~\ref{fig:likert-confidence-after-comp} for Group~A; 5~being the highest confidence).

\section{Conclusion}
\label{sec:conclusion}
Our new CAM system helps users to faster and more confidently find answers on comparative questions compared to a keyword-based search. Moreover, a summary provided in the answer serves to support a decision-making process. While the objects of comparison and the important aspects have still to be stated explicitly, this gives rise to comparative question handling in search engines once respective questions can be identified automatically. A demo of our CAM~system is online\footnote{\url{http://ltdemos.informatik.uni-hamburg.de/cam/}} and available as open source.\footnote{\url{https://github.com/uhh-lt/cam}}

\begin{acks}
This work has been supported by the Deutsche Forschungsgemeinschaft (DFG) within the project ``Argumentation in Comparative Question Answering (ACQuA)'' (grant BI 1544/7-1 and HA 5851/2-1) that is part of the Priority Program ``Robust Argumentation Machines (RATIO)'' (SPP-1999).

\end{acks}

\bibliographystyle{ACM-Reference-Format}
%\bibliography{biblio.bib} 
\bibliography{dblpBiblio.bib}


\begin{thebibliography}{19}

%%% the Web address.

\ifx \showCODEN    \undefined \def \showCODEN     #1{\unskip}     \fi
\ifx \showDOI      \undefined \def \showDOI       #1{#1}\fi
\ifx \showISBNx    \undefined \def \showISBNx     #1{\unskip}     \fi
\ifx \showISBNxiii \undefined \def \showISBNxiii  #1{\unskip}     \fi
\ifx \showISSN     \undefined \def \showISSN      #1{\unskip}     \fi
\ifx \showLCCN     \undefined \def \showLCCN      #1{\unskip}     \fi
\ifx \shownote     \undefined \def \shownote      #1{#1}          \fi
\ifx \showarticletitle \undefined \def \showarticletitle #1{#1}   \fi
\ifx \showURL      \undefined \def \showURL       {\relax}        \fi
% invisible to TeX
\providecommand\bibfield[2]{#2}
\providecommand\bibinfo[2]{#2}
\providecommand\natexlab[1]{#1}
\providecommand\showeprint[2][]{arXiv:#2}

\bibitem[\protect\citeauthoryear{Aker, Sliwa, Ma, Lui, Borad, Ziyaei, and
  Ghobadi}{Aker et~al\mbox{.}}{2017}]%
        {Aker2017What-works-and-}
\bibfield{author}{\bibinfo{person}{Ahmet Aker}, \bibinfo{person}{Alfred Sliwa},
  \bibinfo{person}{Yuan Ma}, \bibinfo{person}{Ruishen Lui},
  \bibinfo{person}{Niravkumar Borad}, \bibinfo{person}{Seyedeh Ziyaei}, {and}
  \bibinfo{person}{Mina Ghobadi}.} \bibinfo{year}{2017}\natexlab{}.
\newblock \showarticletitle{What works and what does not: Classifier and
  feature analysis for argument mining}. In
  \bibinfo{booktitle}{\emph{Proceedings of ArgMining@EMNLP~2017}}.
  \bibinfo{pages}{91--96}.
\newblock


\bibitem[\protect\citeauthoryear{Chen and Guestrin}{Chen and Guestrin}{2016}]%
        {chen2016xgboost}
\bibfield{author}{\bibinfo{person}{Tianqi Chen} {and} \bibinfo{person}{Carlos
  Guestrin}.} \bibinfo{year}{2016}\natexlab{}.
\newblock \showarticletitle{{XGBoost}: {A} scalable tree boosting system}. In
  \bibinfo{booktitle}{\emph{Proceedings of KDD~2016}}.
  \bibinfo{pages}{785--794}.
\newblock


\bibitem[\protect\citeauthoryear{Daxenberger, Eger, Habernal, Stab, and
  Gurevych}{Daxenberger et~al\mbox{.}}{2017}]%
        {Daxenberger2017What-is-the-Ess}
\bibfield{author}{\bibinfo{person}{Johannes Daxenberger},
  \bibinfo{person}{Steffen Eger}, \bibinfo{person}{Ivan Habernal},
  \bibinfo{person}{Christian Stab}, {and} \bibinfo{person}{Iryna Gurevych}.}
  \bibinfo{year}{2017}\natexlab{}.
\newblock \showarticletitle{What is the essence of a claim? Cross-domain claim
  identification}. In \bibinfo{booktitle}{\emph{Proceedings of EMNLP~2017}}.
  \bibinfo{pages}{2055--2066}.
\newblock


\bibitem[\protect\citeauthoryear{Dusmanu, Cabrio, and Villata}{Dusmanu
  et~al\mbox{.}}{2017}]%
        {Dusmanu2017Argument-Mining}
\bibfield{author}{\bibinfo{person}{Mihai Dusmanu}, \bibinfo{person}{Elena
  Cabrio}, {and} \bibinfo{person}{Serena Villata}.}
  \bibinfo{year}{2017}\natexlab{}.
\newblock \showarticletitle{Argument mining on Twitter: Arguments, facts and
  sources}. In \bibinfo{booktitle}{\emph{Proceedings of EMNLP~2017}}.
  \bibinfo{pages}{2317--2322}.
\newblock


\bibitem[\protect\citeauthoryear{Faul, Erdfelder, Lang, and Buchner}{Faul
  et~al\mbox{.}}{2007}]%
        {faul2007g}
\bibfield{author}{\bibinfo{person}{Franz Faul}, \bibinfo{person}{Edgar
  Erdfelder}, \bibinfo{person}{Albert-Georg Lang}, {and} \bibinfo{person}{Axel
  Buchner}.} \bibinfo{year}{2007}\natexlab{}.
\newblock \showarticletitle{G* Power 3: A flexible statistical power analysis
  program for the social, behavioral, and biomedical sciences}.
\newblock \bibinfo{journal}{\emph{Behavior Research Methods}}
  \bibinfo{volume}{39}, \bibinfo{number}{2} (\bibinfo{year}{2007}),
  \bibinfo{pages}{175--191}.
\newblock


\bibitem[\protect\citeauthoryear{Ferrucci, Brown, Chu{-}Carroll, Fan, Gondek,
  Kalyanpur, Lally, Murdock, Nyberg, Prager, Schlaefer, and Welty}{Ferrucci
  et~al\mbox{.}}{2010}]%
        {ferrucci2010building}
\bibfield{author}{\bibinfo{person}{David~A. Ferrucci}, \bibinfo{person}{Eric~W.
  Brown}, \bibinfo{person}{Jennifer Chu{-}Carroll}, \bibinfo{person}{James
  Fan}, \bibinfo{person}{David Gondek}, \bibinfo{person}{Aditya Kalyanpur},
  \bibinfo{person}{Adam Lally}, \bibinfo{person}{J.~William Murdock},
  \bibinfo{person}{Eric Nyberg}, \bibinfo{person}{John~M. Prager},
  \bibinfo{person}{Nico Schlaefer}, {and} \bibinfo{person}{Christopher~A.
  Welty}.} \bibinfo{year}{2010}\natexlab{}.
\newblock \showarticletitle{Building {W}atson: An overview of the {DeepQA}
  project}.
\newblock \bibinfo{journal}{\emph{{AI} Magazine}} \bibinfo{volume}{31},
  \bibinfo{number}{3} (\bibinfo{year}{2010}), \bibinfo{pages}{59--79}.
\newblock


\bibitem[\protect\citeauthoryear{Fiszman, Demner{-}Fushman, Lang, Goetz, and
  Rindflesch}{Fiszman et~al\mbox{.}}{2007}]%
        {fiszman2007interpreting}
\bibfield{author}{\bibinfo{person}{Marcelo Fiszman}, \bibinfo{person}{Dina
  Demner{-}Fushman}, \bibinfo{person}{Fran{\c{c}}ois{-}Michel Lang},
  \bibinfo{person}{Philip Goetz}, {and} \bibinfo{person}{Thomas~C.
  Rindflesch}.} \bibinfo{year}{2007}\natexlab{}.
\newblock \showarticletitle{Interpreting comparative constructions in
  biomedical text}. In \bibinfo{booktitle}{\emph{Proceedings of
  BioNLP@ACL~2007}}. \bibinfo{pages}{137--144}.
\newblock


\bibitem[\protect\citeauthoryear{Franzek, Panchenko, and Biemann}{Franzek
  et~al\mbox{.}}{2018}]%
        {Franzek.9172018}
\bibfield{author}{\bibinfo{person}{Mirco Franzek}, \bibinfo{person}{Alexander
  Panchenko}, {and} \bibinfo{person}{Chris Biemann}.}
  \bibinfo{year}{2018}\natexlab{}.
\newblock \showarticletitle{Categorization of comparative sentences for
  argument mining}.
\newblock \bibinfo{journal}{\emph{CoRR}}  \bibinfo{volume}{abs/1809.06152}
  (\bibinfo{year}{2018}).
\newblock
\urldef\tempurl%
\url{http://arxiv.org/abs/1809.06152}
\showURL{%
\tempurl}


\bibitem[\protect\citeauthoryear{Gupta, Mahmood, Ross, Wu, and
  Vijay{-}Shanker}{Gupta et~al\mbox{.}}{2017}]%
        {gupta2017identifying}
\bibfield{author}{\bibinfo{person}{Samir Gupta}, \bibinfo{person}{A.~S.
  M.~Ashique Mahmood}, \bibinfo{person}{Karen Ross}, \bibinfo{person}{Cathy~H.
  Wu}, {and} \bibinfo{person}{K. Vijay{-}Shanker}.}
  \bibinfo{year}{2017}\natexlab{}.
\newblock \showarticletitle{Identifying comparative structures in biomedical
  text}. In \bibinfo{booktitle}{\emph{Proceedings of BioNLP@ACL~2017}}.
  \bibinfo{pages}{206--215}.
\newblock


\bibitem[\protect\citeauthoryear{Hua and Wang}{Hua and Wang}{2017}]%
        {hua-wang:2017:Short}
\bibfield{author}{\bibinfo{person}{Xinyu Hua} {and} \bibinfo{person}{Lu Wang}.}
  \bibinfo{year}{2017}\natexlab{}.
\newblock \showarticletitle{Understanding and detecting supporting arguments of
  diverse types}. In \bibinfo{booktitle}{\emph{Proceedings of ACL~2017 (Volume
  2: Short Papers)}}. \bibinfo{pages}{203--208}.
\newblock


\bibitem[\protect\citeauthoryear{Jindal and Liu}{Jindal and Liu}{2006}]%
        {JindalLiu2006}
\bibfield{author}{\bibinfo{person}{Nitin Jindal} {and} \bibinfo{person}{Bing
  Liu}.} \bibinfo{year}{2006}\natexlab{}.
\newblock \showarticletitle{Mining comparative sentences and relations}. In
  \bibinfo{booktitle}{\emph{Proceedings of AAAI~2006}}.
  \bibinfo{pages}{1331--1336}.
\newblock


\bibitem[\protect\citeauthoryear{Panchenko, Ruppert, Faralli, Ponzetto, and
  Biemann}{Panchenko et~al\mbox{.}}{2018}]%
        {PANCHENKO18.215}
\bibfield{author}{\bibinfo{person}{Alexander Panchenko}, \bibinfo{person}{Eugen
  Ruppert}, \bibinfo{person}{Stefano Faralli}, \bibinfo{person}{Simone~Paolo
  Ponzetto}, {and} \bibinfo{person}{Chris Biemann}.}
  \bibinfo{year}{2018}\natexlab{}.
\newblock \showarticletitle{Building a web-scale dependency-parsed corpus from
  {CommonCrawl}}. In \bibinfo{booktitle}{\emph{Proceedings of LREC~2018}}.
\newblock


\bibitem[\protect\citeauthoryear{Park and Blake}{Park and Blake}{2012}]%
        {park2012identifying}
\bibfield{author}{\bibinfo{person}{Dae~Hoon Park} {and}
  \bibinfo{person}{Catherine Blake}.} \bibinfo{year}{2012}\natexlab{}.
\newblock \showarticletitle{Identifying comparative claim sentences in
  full-text scientific articles}. In \bibinfo{booktitle}{\emph{Proceedings of
  DSSD@ACL~2012}}. \bibinfo{pages}{1--9}.
\newblock


\bibitem[\protect\citeauthoryear{Shapiro and Wilk}{Shapiro and Wilk}{1965}]%
        {shapiro1965analysis}
\bibfield{author}{\bibinfo{person}{Samuel~Sanford Shapiro} {and}
  \bibinfo{person}{Martin~B. Wilk}.} \bibinfo{year}{1965}\natexlab{}.
\newblock \showarticletitle{An analysis of variance test for normality
  (complete samples)}.
\newblock \bibinfo{journal}{\emph{Biometrika}} \bibinfo{volume}{52},
  \bibinfo{number}{3/4} (\bibinfo{year}{1965}), \bibinfo{pages}{591--611}.
\newblock


\bibitem[\protect\citeauthoryear{Snajder}{Snajder}{2017}]%
        {Snajder2017Social-Media-Ar}
\bibfield{author}{\bibinfo{person}{Jan Snajder}.}
  \bibinfo{year}{2017}\natexlab{}.
\newblock \showarticletitle{Social media argumentation mining: The quest for
  deliberateness in raucousness}.
\newblock \bibinfo{journal}{\emph{CoRR}}  \bibinfo{volume}{abs/1701.00168}
  (\bibinfo{year}{2017}).
\newblock
\urldef\tempurl%
\url{http://arxiv.org/abs/1701.00168}
\showURL{%
\tempurl}


\bibitem[\protect\citeauthoryear{Stab, Daxenberger, Stahlhut, Miller, Schiller,
  Tauchmann, Eger, and Gurevych}{Stab et~al\mbox{.}}{2018}]%
        {stab2018argumentext}
\bibfield{author}{\bibinfo{person}{Christian Stab}, \bibinfo{person}{Johannes
  Daxenberger}, \bibinfo{person}{Chris Stahlhut}, \bibinfo{person}{Tristan
  Miller}, \bibinfo{person}{Benjamin Schiller}, \bibinfo{person}{Christopher
  Tauchmann}, \bibinfo{person}{Steffen Eger}, {and} \bibinfo{person}{Iryna
  Gurevych}.} \bibinfo{year}{2018}\natexlab{}.
\newblock \showarticletitle{{ArgumenText}: Searching for arguments in
  heterogeneous sources}. In \bibinfo{booktitle}{\emph{Proceedings of
  NAACL~2018 (Demonstrations)}}. \bibinfo{pages}{21--25}.
\newblock


\bibitem[\protect\citeauthoryear{Stab and Gurevych}{Stab and Gurevych}{2014}]%
        {Stab2014Identifying-Arg}
\bibfield{author}{\bibinfo{person}{Christian Stab} {and} \bibinfo{person}{Iryna
  Gurevych}.} \bibinfo{year}{2014}\natexlab{}.
\newblock \showarticletitle{Identifying argumentative discourse structures in
  persuasive essays}. In \bibinfo{booktitle}{\emph{Proceedings of EMNLP~2014}}.
  \bibinfo{pages}{46--56}.
\newblock


\bibitem[\protect\citeauthoryear{Sun, Wang, Shen, Zeng, and Chen}{Sun
  et~al\mbox{.}}{2006}]%
        {sun2006cws}
\bibfield{author}{\bibinfo{person}{Jian{-}Tao Sun}, \bibinfo{person}{Xuanhui
  Wang}, \bibinfo{person}{Dou Shen}, \bibinfo{person}{Hua{-}Jun Zeng}, {and}
  \bibinfo{person}{Zheng Chen}.} \bibinfo{year}{2006}\natexlab{}.
\newblock \showarticletitle{{CWS}: A comparative web search system}. In
  \bibinfo{booktitle}{\emph{Proceedings of WWW~2006}}.
  \bibinfo{pages}{467--476}.
\newblock


\bibitem[\protect\citeauthoryear{Wachsmuth, Potthast, Khatib, Ajjour,
  Puschmann, Qu, Dorsch, Morari, Bevendorff, and Stein}{Wachsmuth
  et~al\mbox{.}}{2017}]%
        {wachsmuth2017building}
\bibfield{author}{\bibinfo{person}{Henning Wachsmuth}, \bibinfo{person}{Martin
  Potthast}, \bibinfo{person}{Khalid~Al Khatib}, \bibinfo{person}{Yamen
  Ajjour}, \bibinfo{person}{Jana Puschmann}, \bibinfo{person}{Jiani Qu},
  \bibinfo{person}{Jonas Dorsch}, \bibinfo{person}{Viorel Morari},
  \bibinfo{person}{Janek Bevendorff}, {and} \bibinfo{person}{Benno Stein}.}
  \bibinfo{year}{2017}\natexlab{}.
\newblock \showarticletitle{Building an argument search engine for the web}. In
  \bibinfo{booktitle}{\emph{Proceedings of ArgMining@EMNLP~2017}}.
  \bibinfo{pages}{49--59}.
\newblock


\end{thebibliography}
\end{document}